%% file: main.tex
\documentclass[runningheads]{llncs}
\usepackage[T1]{fontenc}

\usepackage{graphicx}
\usepackage{hyperref}
\usepackage{url}
\usepackage{booktabs}
\usepackage{amsfonts}
\usepackage{nicefrac}
\usepackage{microtype}
\usepackage{xcolor}
\usepackage{amsmath}
\usepackage{makecell}
\usepackage{graphicx}
\usepackage{multirow}
\usepackage{relsize}
\usepackage[figuresleft]{rotating}
\usepackage{tabularx} 
\usepackage{float}
\begin{document}

\title{\texttt{RAFEN} -- Regularized Alignment Framework for Embeddings of Nodes}


\author{
Kamil Tagowski\inst{1}\orcidID{0000-0003-4809-3587} \and
Piotr Bielak\inst{1}\orcidID{0000-0002-1487-2569} \and
Jakub Binkowski\inst{1}\orcidID{0000-0001-7386-5150} \and
Tomasz Kajdanowicz\inst{1}\orcidID{0000-0002-8417-1012}
}

\authorrunning{K. Tagowski et al.}

\institute{
Department of Artificial Intelligence,\\
Wrocław University of Science and Technology,\\
Wrocław, Poland\\
\email{kamil.tagowski@pwr.edu.pl}
}

\maketitle

\begin{abstract}
Learning representations of nodes has been a crucial area of the graph machine learning research area. A well-defined node embedding model should reflect both node features and the graph structure in the final embedding. In the case of dynamic graphs, this problem becomes even more complex as both features and structure may change over time. The embeddings of particular nodes should remain comparable during the evolution of the graph, what can be achieved by applying an alignment procedure. This step was often applied in existing works after the node embedding was already computed. In this paper, we introduce a framework -- \texttt{RAFEN} -- that allows to enrich any existing node embedding method using the aforementioned alignment term and learning aligned node embedding during training time. We propose several variants of our framework and demonstrate its performance on six real-world datasets. \texttt{RAFEN} achieves on-par or better performance than existing approaches without requiring additional processing steps.

\keywords{Dynamic Graphs \and Graph Embedding \and Embedding Alignment \and Graph Neural Networks \and Link prediction  \and Machine Learning.}
\end{abstract}

\section{Introduction}

In recent years, representation learning on structured objects, like graphs, gained much attention in the community. The wide application range, which includes recommender systems, molecular biology, or social networks, motivated a rapid development of new methods for learning adequate representations of such data structures. Especially real-world data often involves accounting for temporal phenomena, which requires handling structural changes over time (i.e., dynamic graphs).

Researchers widely use the embedding of dynamic graphs in a discrete-time manner, i.e., the graph data is aggregated into snapshots and embeddings are computed for each one of them. In opposite to the continuous-time approach, one does not require to provide an online update mechanism, thus allowing one to utilize any classical static embedding approaches or temporal embedding. On the other hand, the discrete-time approach requires an embedding aggregation method that transforms the sequence of node snapshot embeddings into a single node embedding matrix which captures the whole graph evolution. Unfortunately, due to the stochastic properties of node embedding methods, such embeddings are algebraically incomparable. To solve this problem, most existing approaches apply a post-hoc correction \cite{bielak2022fildne,singer2019node,tagowski2021alingment,trivedi2018structural}. 

In this work, we consider a different scenario where we build aligned node embeddings by introducing an additional regularization term to a model's loss function. This term explicitly aligns embeddings of corresponding nodes from consecutive timesteps, which eventually helps to improve results on a downstream task. In particular, we summarize the contributions of the paper as follows:
\begin{enumerate}
    \item[C1.] We propose a novel framework \texttt{RAFEN} for learning aligned node embeddings in dynamic networks. It can be used with any existing node embedding method (both for static and dynamic networks) and, contrary to existing post-hoc methods, it does not rely on a time-consuming matrix factorization approach.
    \item[C2.] We propose three different versions of the alignment function, including one that treats all nodes equally and one that utilizes temporal network measures for weighting the alignment process. 
    \item[C3.] We evaluate the proposed framework variants in a link prediction task over six real-world datasets and show that our approach allows to improve results on the downstream task compared to other existing methods.
\end{enumerate}

\section{Related Work}

\paragraph{\textbf{Static Node Embedding.}} To date, plethora of node embedding methods have been introduced. The seminal work of Grover et al. \cite{grover2016node2vec} introduced Node2Vec method, which enables learning structural embeddings of nodes in an unsupervised manner. It pushes node embeddings apart when nodes are away from each other but makes embeddings closer for nearby nodes, leveraging nodes co-occurrences on common random walks as a measure of node "distance". Recently, the main body of literature was focused on Graph Neural Networks, among which GCN \cite{kipf_semi-supervised_2017}, GIN \cite{xu_how_2019}, and GAT \cite{velivckovic2017graph} are the most notable ones. In particular, the Graph AutoEncoder (GAE) \cite{kipf2016variational} architecture leverages GCNs to learn unsupervised node structural embeddings, using reconstruction of the adjacency matrix as the training task. As static node embeddings can be adopted for time-discrete dynamic graphs, we utilize Node2Vec and GAE as base for computing dynamic node representations in this work.

\paragraph{\textbf{Dynamic Node Embedding.}} The problem of dynamic node embedding has recently gained more focus from researchers. We can distinguish meta approaches that could utilize any static embedding method to produce dynamic snapshot embeddings \cite{bielak2022fildne,singer2019node,tagowski2021alingment,trivedi2018structural}. However, most of them utilize only the Node2Vec \cite{grover2016node2vec} embeddings method. Another group of models are the ones dedicated for dynamic node embedding, such as Dyngraph2vec \cite{goyal2020dyngraph2vec}, where the authors utilize the Autoencoder (AE) architecture to capture the temporal network's evolving structure and provide three variants of the model which differ in the representation method of a neighbor vector.  

\paragraph{\textbf{Learning of compatible representation.}} This problem has already been studied in literature \cite{hu2022learning,meng2021learning,shen2020towards}. In \cite{hu2022learning}, the authors present the BC-Aligner method that allows obtaining previous embedding models based on a learnable transformation matrix.  \cite{meng2021learning} achieves model compatibility with an alignment loss that aligns class centers between models with a boundary loss to constraint new features to be more centralized to class centers. \cite{shen2020towards} proposes backward compatibility training for image classification by adding an influence loss that can capture dataset changes along with new classes.

\paragraph{\textbf{Post-hoc alignment for node embeddings}} There are already approaches for alignment of node embeddings based on the Orthogonal Procrustes method  \cite{bielak2022fildne,singer2019node,tagowski2021alingment,trivedi2018structural}. In \cite{singer2019node,trivedi2018structural}, the authors use all common nodes between snapshots to perform the alignment. In \cite{bielak2022fildne}, it has been found that some nodes change their behavior too significantly. Therefore, they use only a subset of common nodes, which is selected based on activity and selection method. This approach in further explored in \cite{tagowski2021alingment} by introducing a wide range of activity functions that can be used for the alignment process.

\section{The proposed \texttt{RAFEN} framework}\label{sec:rafen}

\begin{definition}
\label{def:dyngraph}
A dynamic network (graph) is a tuple $G_{0, T} = (V_{0,T}, E_{0, T})$, where $V_{0, T}$ denotes a set of all nodes (vertices) observed between timestamp $0$ and $T$, and $V_{0,T}$ is a set of edges in the same timestamp range. We model such a network as a sequence of graph snapshots $G_{0,1}, G_{1,2}, G_{2,3}, \ldots, G_{T-1, T}$.
\end{definition}

\begin{definition}
Node embedding is a function $f: V \to {\mathbb{R}^{|V| \times d}}$ that maps a set of nodes $V$ into low-dimensional (i.e., $d \ll |V|$) embedding matrix $F$, where each row represents the embedding of a single node.
\end{definition}

\begin{definition}
\label{def:scoring}

Node activity scoring function $s: V_{com} \to \mathbb{R}^{|V_{com}|}$ (with $V_{com} = V_{i, j} \cap V_{k, l}$ and $i<j<k<l$) assigns a scalar score to each node from the set of common nodes $V_{com}$, which reflects the change in a node's activity between two snapshots $G_{i, j}$ and $G_{k, l}$.
\end{definition}

\begin{definition}
Reference nodes $V_{ref}$ is a subset of common nodes between two snapshots retrieved by a selection function. We use the percent selection function as defined in \cite{bielak2022fildne}, with chooses the top $p$\% of nodes -- i.e., $select(S, V) = V_{ref} \subseteq \text{sort}_{S}(V), \ \text{s.t.} \ |V_{ref}| = \textbf{p}|V|$. 
\end{definition}

\paragraph{\textbf{Problem statement.}} Given a dynamic graph $G_{0,T}$ in the form of discrete graph snapshots (see Definition \ref{def:dyngraph}), the objective is to find at any given timestamp $t \in \{1, 2, \ldots, T\}$, the node embedding $F_{t-1,t}$ of the current snapshot $G_{t-1, t}$, such that $F_{t-1,t}$ will be algebraically compatible (aligned) with the node embedding $F_{t-2,t-1}$ of the previous snapshot $G_{t-2, t-1}$. Such compatibility is required to properly aggregate snapshot embeddings $F_{0,1}, F_{1,2}, \ldots, F_{t-1, t}$ into a single node embedding matrix $F_{0, t}$ that summarizes the whole graph evolution (see \cite{bielak2022fildne} for more details about the aggregation mechanism).

\begin{figure}[!ht]
    \centering

    \includegraphics[scale=1.08]{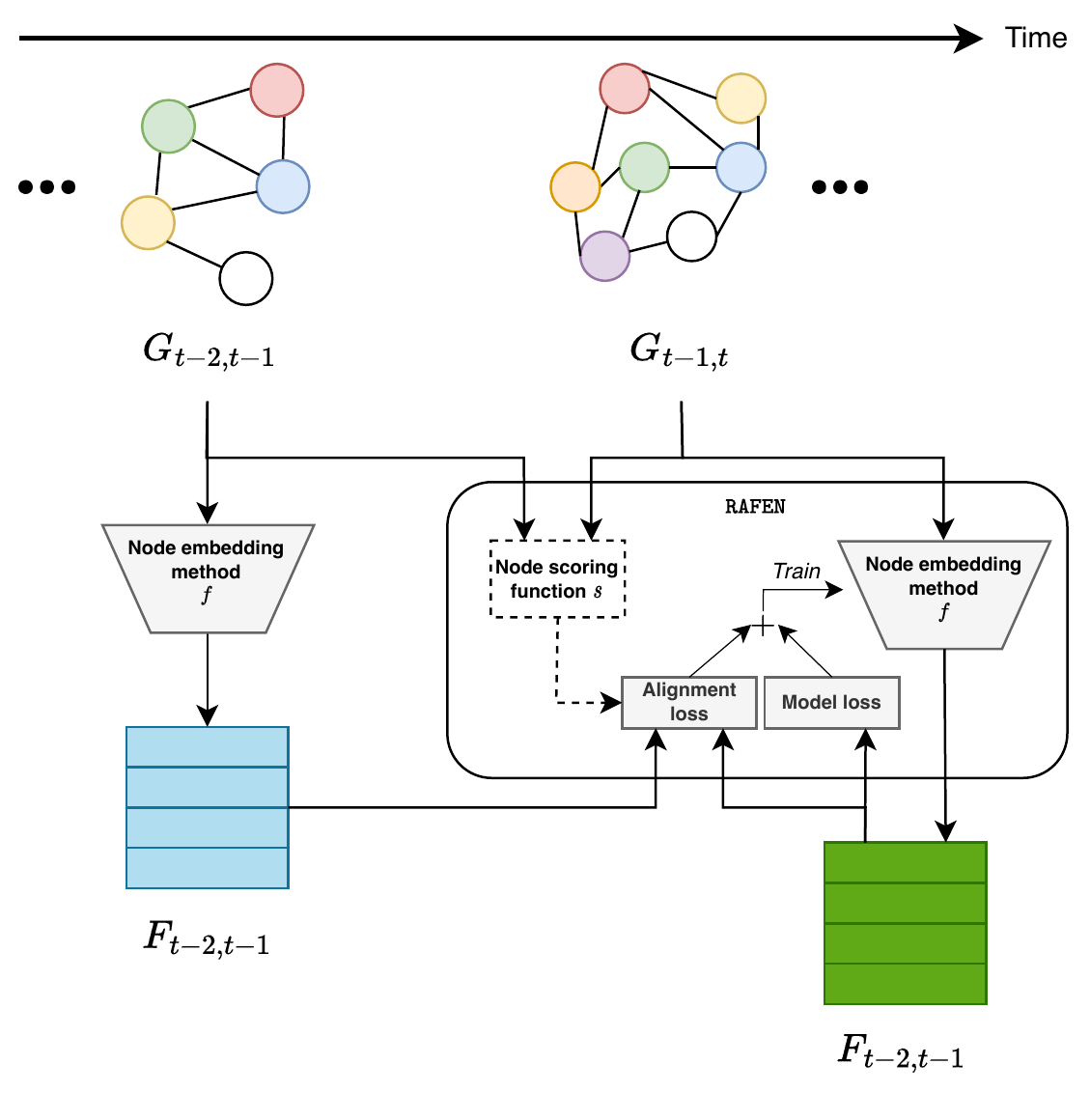}
        \caption{
     Overview of the \texttt{RAFEN} pipeline. The pipeline can be decomposed into two steps: (1) For the snapshot $G_{t-2, t-1}$ that we want align to, we compute node embeddings  $F_{t-2, t-1}$ using the vanilla node embedding method $f$. (2) For the next snapshot $G_{t-1,t}$, we additionally compute the alignment loss term, which we add to the node embedding method model loss -- Eq. \eqref{eq:loss}. The alignment loss term is calculated as an MSE (mean squared error) -- Eq. \eqref{eq:alignment_loss} between the representation of common nodes of currently trained node embeddings $F_t{-1, t}$ and embeddings from the previous snapshot $F_{t-2, t}$. There is no need to take all common nodes to calculate the alignment loss. Instead, we apply node scoring function $s$ that computes the difference in node's activity between snapshots and selects a subset of common nodes -- Eq. \eqref{eq:alignment_loss_vref} or use the output as weights in alignment loss term -- Eq. \eqref{eq:weighted_alignment_loss}}
    \label{fig:pipeline}
\end{figure}

\paragraph{\textbf{RAFEN.}} As a solution for the alignment problem, we propose a novel framework \texttt{RAFEN} that enhances the loss function of any existing node embedding method $L_\text{model}$ by means of an alignment regularization term $L_\text{alignment}$. It allows to learn node embeddings that are aligned with a given anchor embedding (in our case: the previous graph snapshot). Our framework can be used along with expert knowledge by setting a model hyperparameter $\alpha$ or using temporal network measures to automatically determine the alignment coefficient of each node. We will discuss all framework details in the sections below. Moreover, we present a general overview of our pipeline in Figure \ref{fig:pipeline}.
 
\paragraph{\textbf{How to incorporate the alignment term?}} We enhance the loss function of the node embedding model $L_{model}$ by adding an alignment term $L_{alignment}$. In order to balance between the information from the previous graph snapshot (through the optimization of the $L_{alignment}$ loss) and the information from the current snapshot (through the optimization of the $L_{model}$ loss), we propose to combine both terms in the following way: 

\begin{equation}
\label{eq:alpha_loss}
    L = (1-\alpha) \cdot L_{model} + \alpha \cdot L_{alignment}
\end{equation}

The $\alpha$ parameter controls the trade-off between the model's loss and the alignment term, which are both critical factors in the model's learning behavior.
We can distinguish two boundary conditions of the $\alpha$ parameter:
\begin{itemize}
    \item $\alpha = 0$ -- the model uses only model loss,
    \item $\alpha = 1$ -- the model uses only alignment term.
\end{itemize}

Choosing the right $\alpha$ value is essential for the model's performance. It can be done by a simple grid search or through expert knowledge. However, performing an $\alpha$ grid search might be time-consuming and computationally expensive. 
To address this, we propose a simplified version of our framework, where we set equal importance to both loss terms, which is equivalent to setting $\alpha = 0.5$

\begin{equation}
\label{eq:loss}
    L = L_{model} + L_{alignment}
\end{equation}

\paragraph{\textbf{Alignment loss term ($L_{alignment}$)}} The foundation of our proposed \texttt{RAFEN} framework is the alignment loss term $L_{\text{alignment}}$. We build upon the mean squared error (MSE) between node $v$'s embedding from the current snapshot $F_{t-1, t}^{(v)}$ and its representation from the previous snapshot $F_{t-2,t-1}^{(v)}$. We apply such regularization to a subset of the common nodes $V_{com} \in V_{t-2,t-1} \cap V_{t-1, t}$.
\clearpage
Let us now define three \texttt{RAFEN} variants:

\begin{itemize}
    \item \texttt{RAFEN\_ALL} -- we use all common nodes between consecutive snapshots:

    \begin{equation}
    \label{eq:alignment_loss}
    L_{\text{alignment}} = \frac{1}{|V_{com}|} \mathlarger{\sum_{v \in V_{com}}} \left(F_{t-1, t}^{(v)} - F_{t-2, t-1}^{(v)} \right)^2
    \end{equation}

    \item \texttt{RAFEN\_Weighted} -- similarly to the previous case, we use all common nodes, but instead of treating each one of them equally, we assign weights that reflect the node activity change (node activity scoring function $s(\cdot)$ - see Definition \ref{def:scoring}) of a particular node between the graph snapshots;

    \begin{equation}
    \label{eq:weighted_alignment_loss}
    L_{\text{alignment}} = \frac{1}{|V_{com}|} \mathlarger{\sum_{v \in V_{com}}} \left(\left(F_{t-1, t}^{(v)} - F_{t-2, t-1}^{(v)} \right)^2 \cdot s(v) \right)
    \end{equation}

    \item \texttt{RAFEN\_REF} -- in \cite{bielak2022fildne,tagowski2021alingment}, the authors show that utilizing all common nodes in the alignment process may lead to degraded performance in the downstream tasks due to the fact that some nodes change their behavior too much. They use $V_{ref} \subset V_{com}$ subset of common nodes, which is built according to the node activity scores (the top $p$\% of the common nodes are selected).

    \begin{equation}
    \label{eq:alignment_loss_vref}
    L_{\text{alignment}} = \frac{1}{{|V_\text{ref}|}} \mathlarger{\sum_{v \in V_\text{ref}}} \left(F_{t-1, t}^{(v)} - F_{t-2, t-1}^{(v)}\right)^2
    \end{equation}
\end{itemize}

\section{Experimental Setup}
We evaluate all methods for the link prediction task. Following, we discuss the details and hyperparameters of the experiments.

\subsection{Datasets}
We conducted experiments on six real-world datasets. We followed the graph snapshots split procedure defined in \cite{tagowski2021alingment}, such that we split graphs based on the timestamp frequency (monthly or yearly). Moreover, the first four snapshots in \textit{ppi}, and \textit{ogbl-collab} datasets were ignored, as merging them would result in a too broad timespan of such merged snapshots. Also, the two last snapshots of \textit{bitcoin-alpha} and \textit{bitcoin-otc} were merged to provide more data for validation. We present an overview of datasets in Table \ref{tab:datasets}. Despite the similarities with the experimental setup in \cite{tagowski2021alingment}, we reproduced all experiments due to a refined evaluation protocol on link prediction. Also, our method differs in alignment procedure, i.e., it aligns embedding to the previous snapshot instead of the first.

\input{tables/datasets}

\subsection{Node embeddings}
We selected two different methods as our base models for computing node embeddings. First, we used Node2Vec, a widely renowned node structural embedding method that relies on random walks. To show our framework's universality, we included a second method -- GAE, an encoder-decoder model that utilizes the graph convolution operation. Both methods are capable of capturing meaningful structural features of the graph and are trained in an unsupervised manner. Nonetheless, due to the flexibility of our framework, other node embedding models could also be utilized. For both models, each snapshot representation was recomputed 25 times to account for randomness in these methods (e.g., random walks in Node2Vec, weight initialization in the encoder of the GAE model). 

\paragraph{\textbf{Node2Vec}} We trained the Node2Vec model, reusing the hyperparameters configuration from \cite{tagowski2021alingment}, as these were obtained from a hyperparameter search procedure. In particular, we use 128-dimensional embeddings, except for the \textit{bitcoin-alpha} dataset where we use 32 as the embedding size. Further, we use the same configurations in our \textit{RAFEN} models, as we desire that our models perform well without additional hyperparameter search. We utilize the Node2Vec implementation from the PyTorch Geometric library \cite{Fey/Lenssen/2019}.

\paragraph{\textbf{GAE}} We use a Graph Autoencoder with two Graph Convolutional layers (GCN) as an encoder and the Inner-Product decoder. Since datasets do not come with node features required by the GCN layers, we use an additional trainable lookup embedding layer, which serves as a node feature matrix for the model (we evaluated different strategies in preliminary experiment -- this one turned out to work the best). We train these models for $100$ epochs using the Adam optimizer with a 
 learning rate of $0.01$, and a hidden layer size of $128$.

\subsection{Aligned models} 
We tested variants of our \texttt{RAFEN} framework with loss functions introduced in section \ref{sec:rafen}, namely \texttt{RAFEN\_ALL}, \texttt{RAFEN\_WEIGHTED}, and \texttt{RAFEN\_REF}. 
In addition, \texttt{RAFEN\_WEIGHTED} involves scoring function, denoted as $s(\cdot)$ in Eq. \eqref{eq:weighted_alignment_loss}. For scoring function, we selected \textit{Edge Jaccard} ($\text{EJ}$) \cite{tagowski2021alingment}, and \textit{Temporal Betweenness} ($\text{TB}$) \cite{yu2020identifying} due to their good performance in post-hoc alignment methods in previous work. Also, these two functions provide desired variability, such that they are from a different family of methods: \textit{Edge Jaccard} is based only on edge changes in the static scheme, while \textit{Temporal Betweenness} takes into account the time aspect. We denote the two variants as \texttt{RAFEN\_WEIGHTED\_EJ} and \texttt{RAFEN\_WEIGHTED\_TB}, for \textit{Edge Jaccard} and \textit{Temporal Betweenness} respectively.

It is worth noting that $L_{model}$ and $L_{alignment}$ are in different scales. Therefore, we additionally scale them with the loss value of the model from the first batch.

\subsection{Link Prediction}
We define the link prediction task as edge existence prediction on the last snapshot of the graph $G_{T-1, T}$, based on the previously learned representations from previous snapshots $F_{0,1}, \ldots, F_{T-2, T-1}$. We take existing links in the graph $G_{T-1, T}$ as positive examples. To avoid out-of-distribution nodes in evaluation, we filter edges wherein one of the nodes was not observed previously. The  number of negative edges (non-existing edges) is equal in size to the positive ones. In the negative sampling process, we also applied a simple edge reject criterion that prevents adding already existing edges as negative ones. Having edges sampled, we split the datasets into train, val and test splits in the proportion: 60\%, 20\%, 20\%, respectively. This split is both leveraged during grid-search hyperparameter optimization and the method evaluation. Finally, using optimal hyperparameters, we performed the final evaluation with training and validation subsets merged. The results of the evaluation are reported in Section \ref{sec:results}. 

 To fuse the node embeddings from all snapshots into a single one that reflects the graph evolution, we evaluate 4 different aggregation schemes:
\begin{itemize}
    \item \texttt{mean} -- node embeddings are averaged across snapshots,
    \item \texttt{last} -- only the node embedding from the last snapshot is taken,
    \item \texttt{FILDNE} -- weighted incremental combination of previous snapshot embeddings as in \cite{bielak2022fildne},
    \item \texttt{k-FILDNE} -- weighted incremental combination of previous snapshot embeddings with automated weight estimation as in \cite{bielak2022fildne}.
\end{itemize}

Further, we use a logistic regression classifier, which takes the Hadamard product of the two aggregated node embeddings. We utilize the implementation from the scikit-learn \cite{scikit-learn} library with \textit{liblinear} solver, keeping other hyperparameters values default.

Summarizing, the experiments were performed on link prediction with embeddings trained with 3 variants of loss function and 4 embeddings aggregation schemes. We also report results for the two embedding methods (Node2Vec, GAE) trained without alignment. In addition, we used 3 post-hoc alignment methods from \cite{tagowski2021alingment} as our baselines. To sum up, we evaluate the following models:

\begin{itemize}
    \item \textbf{(whole graph)} -- embeddings are trained on whole graph $G_{0, T-1}$,
    \item \textbf{(last snapshot only)} -- embeddings are trained on last graph snapshot $G_{T-2, T-1}$,
    \item \textbf{Posthoc-PA} -- Procrustes Aligner (PA), 
    \item \textbf{Posthoc-EJ} -- Edge Jaccard Aligner (EJA), 
    \item \textbf{Posthoc-TB} -- Temporal Betweenness (TB),
    \item \texttt{RAFEN\_ALL} -- \texttt{RAFEN} variant with all common nodes between consecutive snapshots and simplified loss importance term,
    \item \texttt{RAFEN\_Weighted\_EJ} -- \texttt{RAFEN\_Weighted} variant with Edge Jaccard node activity scoring method and simplified loss importance term, 
    \item \texttt{RAFEN\_Weighted\_TB} -- \texttt{RAFEN\_Weighted} variant with Temporal Betweenness node activity scoring method and simplified loss importance term.
\end{itemize}

\section{Results}\label{sec:results}

This section presents the results of our experiments, divided into three subsections presenting different framework characteristics. First, we compare the embedding aggregation methods. Next, we show the results for the two formulations of \texttt{RAFEN} loss function, i.e., with and without hyperparameter $\alpha$. Finally, we present the results of several variants of  \texttt{RAFEN} models compared to the baselines approaches.

\subsection{Embeddings aggregation method comparison}\label{sec:aggregation}

Foremost, we evaluated the embedding aggregation methods, namely \texttt{last}, \texttt{mean}, \texttt{FILDNE}, \texttt{k-FILDNE}. We summarize these results in Table \ref{tab:embeddings_aggregation}. We chose the best aggregation method that performs well on both GAE and Node2Vec-based models based on mean ranks for each dataset separately, i.e., bitcoin-alpha: \texttt{k-FILDNE}; bitcoin-otc: \texttt{last}; fb-forum: \texttt{FILDNE}; fb-messages: \texttt{k-FILDNE}; ogbl-collab: \texttt{last}, ppi: \texttt{last}. Please note that for the bitcoin-alpha dataset, the best aggregation method for Node2Vec and GAE differs. However, we chose \texttt{k-FILDNE} due to a better combination of mean ranks (i.e., $1.22$ and $1.56$ vs $3.44$ and $1.44$).

\input{tables/embeddings_agregations}

\begin{figure}[ht!]
    \centering
    \includegraphics[width=\textwidth]{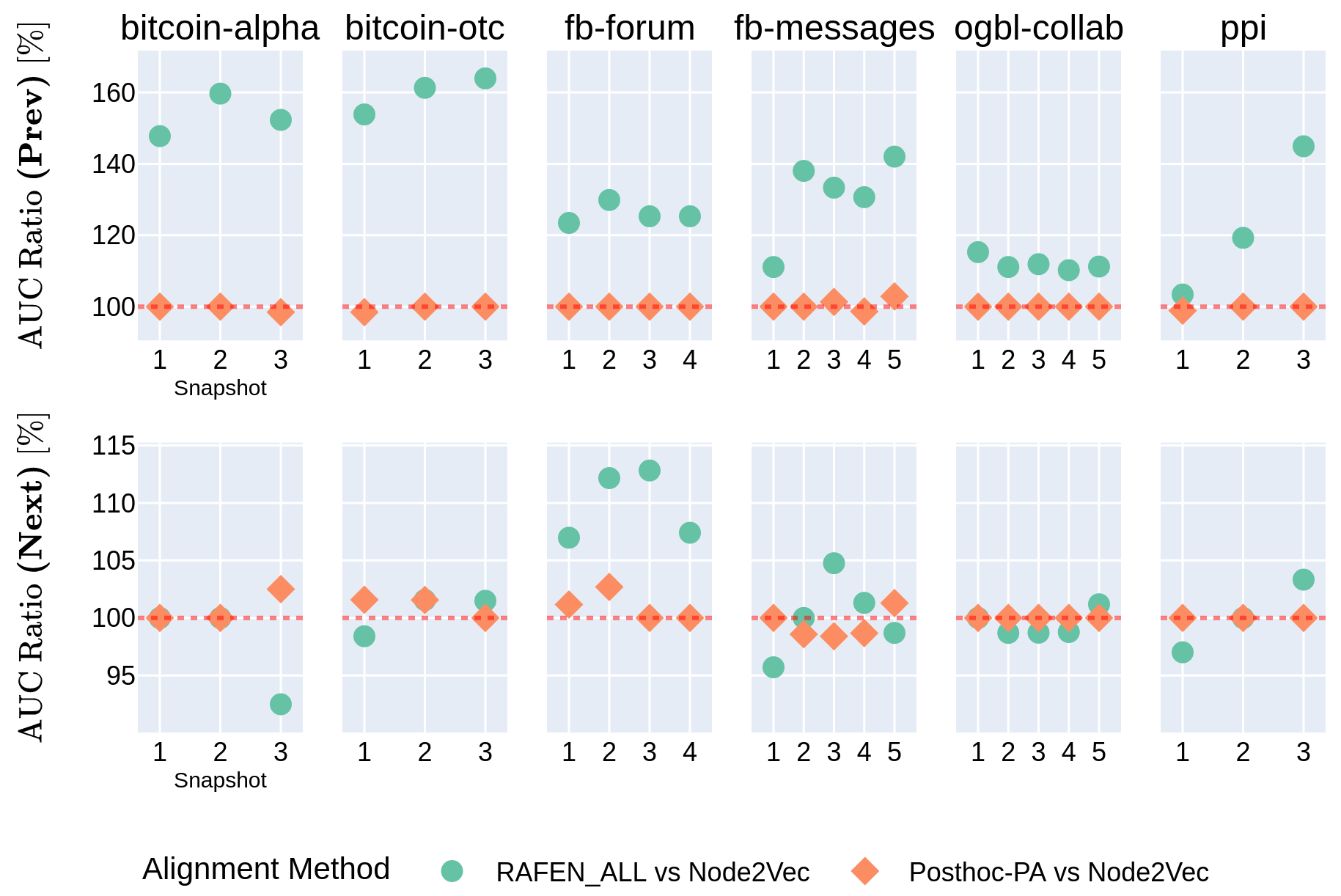}
    \caption{Single snapshot representation evaluation study. In the first figure row, the previous snapshot evaluation is shown (denoted as \textbf{Prev}), and in the second row next snapshot evaluation (denoted as \textbf{Next}). Each column represents a different dataset. Y-Axis contains a link prediction AUC metric Ratio calculated between Alignment Method and vanilla Node2Vec. X-Axis presents snapshot IDs on which node representation was trained.}
    \label{fig:knowledge_transfer}
\end{figure}

Counterintuitively to previous dynamic graph representation learning papers \cite{bielak2022fildne} for four out of six datasets, the best aggregation method was \texttt{last}, which does not aggregate all historical representations. That behavior led us to an additional study that we conducted to further explore both Posthoc and \texttt{RAFEN} methods. We 
 contrasted AUC on the link-prediction of \texttt{RAFEN\_ALL} and \texttt{Posthoc-PA} with plain Node2Vec embeddings and computed their relative performance. In particular, we evaluated two scenarios:
\begin{itemize}
\item \textbf{Prev} -- we take each node's $v$ embedding $F_{t-1, t}^{(v)}$ at time step $t$ and evaluate it on link-prediction using the previous snapshot's graph structure $G_{t-2, t-1}$ as a target,
\item \textbf{Next} --  we take each node's $v$ embedding $F_{t-1, t}^{(v)}$ at time step $t$ and evaluate it on link-prediction using next snapshot's graph structure $G_{t, t+1}$ as a target.
\end{itemize}
We report the link prediction AUC ratio between alignment methods and the non-aligned Node2Vec version. Results are shown in Figure \ref{fig:knowledge_transfer}.

Our proposed \texttt{RAFEN} method increased the evaluation performance on previous snapshots for all of the benchmarked datasets. The gain ranges from 3.5\% on the first snapshot of the ppi dataset up to a 63\% difference on the third snapshot of bitcoin-otc. In the case of the \textbf{Next} scenario, we observe that the last snapshot evaluation performance increased on four our of six datasets -- ranging from 1.2\% in the case of the ogbl-collab dataset up to 7.4\% in fb-forum.

\subsection{Comparison of \texttt{RAFEN}'s $\alpha$-based variants}
\label{sec:rafen_alpha_variants}

In section \ref{sec:rafen}, we proposed two formulations of the loss function for our \texttt{RAFEN} framework. Since the first variant defined in Eq. \eqref{eq:alpha_loss} involves an $\alpha$ hyperparameter, which requires additional domain knowledge or hyperparameter search procedure (e.g., grid search), we investigate whether its simplified variant with $L_{model}$ and $L_{alignment}$ treated equally important provides competitive performance. Likewise, \texttt{RAFEN\_REF} models require defining a reference node selection function with hyperparameters, like top $p$\% in our case. Therefore, we compare it to \texttt{RAFEN\_Weighted} models that rely only on scoring function and do not involve additional hyperparameters. Results of the comparisons are shown in Table \ref{tab:rafen_models_comparison}.

\input{tables/alignment_loss_weight_comparison}

Across almost all the shown comparisons, we can observe that percentage AUC difference in $alpha$-based models to simplified/weighted versions is in the range (-1.61\% to 1.71\%). We can observe outlier values in both \textbf{bitcoin-alpha} and \textbf{ppi} favoring $\alpha$ hyperparameter. Conversely, there is an outlier in \textbf{fb-forum} favoring the weighted method. As in most cases, the differences are negligible, we decided to use simplified/weighted versions for the comparison with baselines. In the aftermath, \texttt{RAFEN} requires only one additional hyperparameter compared to plain GAE or Node2Vec models, which is the node activity scoring function.

\subsection{\texttt{RAFEN} comparison to baselines}\label{sec:rafen_to_baselines}

The results of the comparison between our proposed \texttt{RAFEN} framework and baseline models are presented in Table \ref{tab:baselines_comparison}. Compared to snapshot-based Node2Vec and GAE baselines, \texttt{RAFEN} was better on four datasets: bitcoin-otc, fb-forum, ppi, and ogbl-collab -- only losing to the variant trained on the full graph data. For the remaining two datasets, namely fb-messages, and bitcoin-alpha, we observe the superior performance of post-hoc methods. However, as the previous study (see Section \ref{sec:rafen_alpha_variants}) shows, these datasets provided the most significant differences between $\alpha$ based and weighted. When the models were tuned, they would have also improved performance compared to \texttt{RAFEN\_Weighted} on these datasets.

In the case of \texttt{RAFEN\_Weighted}, the temporal betweenness scoring function was barely better than Edge Jaccard, showing notable differences only in the case of GAE on ogbl-collab and Node2Vec on fb-forum. When comparing \texttt{RAFEN\_ALL} and \texttt{RAFEN\_Weighted}, \texttt{RAFEN\_ALL} approach served as a strong baseline, showing better performance and higher rank in most of the experiments. We hypothesize that the scoring functions we leveraged led to decreased performance. Hence, it requires further investigation and refinements, which we leave for future work.

\input{tables/method_results}

\section{Conclusion and future work}
In this paper, we tackled the problem of learning aligned node embedding for dynamic graphs modeled as a series of discrete graph snapshots. Contrary to existing approaches, such as post-hoc methods, we perform the alignment step during the embedding training. We proposed a novel framework -- \texttt{RAFEN} -- which allows to enrich any existing node embedding method with the aforementioned alignment capabilities. We conducted experiments on six real-world datasets and showed that our approach, even in its simplified version (with minimum set of hyperparameters), achieves better or on-par performance compared to existing approaches. For future work, we want to explore more advanced mechanisms for the actual alignment step, including better investigation of the scoring function in \texttt{RAFEN\_Weighted} setting and the applicability of \texttt{RAFEN} for continuous-time dynamic graphs.

Our code with experiments and data, all enclosed in DVC pipelines \cite{ruslan_kuprieiev}, is available publicly at \url{https://github.com/graphml-lab-pwr/rafen}.

\clearpage

\subsubsection{Acknowledgments} 

This work was financed by (1) the Polish Ministry of Education and Science, CLARIN-PL; (2) the European Regional Development Fund as a part of the 2014–2020 Smart Growth Operational Programme, CLARIN — Common Language Resources and Technology Infrastructure, project no. POIR.04.02.00-00C002/19; (3) the statutory funds of the Department of Artificial Intelligence, Wroclaw University of Science and Technology, Poland; (4) Horizon Europe Framework Programme MSCA Staff Exchanges grant no. 101086321 (OMINO).

\bibliographystyle{splncs04}
\bibliography{main}

\end{document}

%% file: tables/datasets.tex
\begin{table}[ht]
 \caption{
 Statistics of graph datasets. $\mathbf{\mathcal{|V|}}$ - number of nodes, $\mathbf{\mathcal{|E|}}$ - number of edges, Dir - whether the graph is directed or not, T - number of snapshots
 }  
     
    \label{tab:datasets}
    
    \begin{center}
        \scalebox{1.1}{

        \begin{tabular}{lrrcrcrr}
        \toprule
        \textbf{Dataset} & $|\mathbf{\mathcal{V}|}$ & $|\mathbf{\mathcal{E}|}$ & \textbf{Dir} & \textbf{Timespan} & \textbf{T} & \textbf{Snapshot} & \textbf{Network}\\
        & & & & &  & \textbf{timespan} & \textbf{domain}\\
        \midrule
        fb-forum \cite{network_repository} & 899 & 33 720 & $\times$ & 5.5 months & 5 & 1 month & social  \\
        fb-messages \cite{network_repository} & 1 899 & 61 734 & $\times$ & 7.2 months & 7 & 1 month & social \\
        bitcoin-alpha \cite{network_repository} & 3 783 & 24 186 & $\surd$ & 5.2 years & 5 & 1 year & crypto  \\
        bitcoin-otc \cite{network_repository} & 5 881 & 35 592 & $\surd$ & 5.2 years & 5 & 1 year & crypto  \\
        ppi \cite{network_repository} & 16 386 & 141 836 & $\times$ & 24 years & 5 & 5 years & protein  \\
        ogbl-collab \cite{hu2020ogb} & 233 513 & 1 171 947 & $\times$ & 34 years & 7 & 5 years & citation  \\

        \bottomrule
        \end{tabular}
        }
    \end{center}
\end{table}

%% file: tables/embeddings_agregations.tex
\begin{table}[ht]
  \centering
      \caption{Mean ranks of aggregation methods on different datasets. Ranks were calculated on the link prediction task (evaluated on last snapshot), using the AUC metric over 25 runs. Then we took mean ranks of all the \texttt{RAFEN} and Posthoc models for Node2Vec and GAE models, seperately.}
  \label{tab:embeddings_aggregation}
\begin{tabularx}{\textwidth}{XX|*4{>{\centering\arraybackslash}X}@{}}
\toprule
&  & \multicolumn{4}{c}{\textbf{Aggregation method}}\\ 
     \textbf{Dataset}  &  \textbf{Model}   &           \texttt{FILDNE} &         \texttt{k-FILDNE} &             \texttt{last} &  \texttt{mean} \\
\midrule
\multirow[c]{2}{*}{bitcoin-alpha} & Node2Vec & 2.11 & $\mathbf{1.22}$ & 3.22 & 3.44 \\
& GAE & 3.00 & 1.56 & 4.00 & $\mathbf{1.44}$ \\
\midrule
\multirow[c]{2}{*}{bitcoin-otc} & Node2Vec & 1.89 & 3.00 & $\mathbf{1.11}$ & 4.00 \\
& GAE & 1.67 & 2.89 & $\mathbf{1.44}$ & 4.00 \\
\midrule
\multirow[c]{2}{*}{fb-forum} & Node2Vec & $\mathbf{1.22}$ & 2.11 & 3.11 & 3.56 \\
& GAE & $\mathbf{1.33}$ & 2.22 & 2.78 & 3.67 \\
\midrule
\multirow[c]{2}{*}{fb-messages} & Node2Vec & 2.00 & 3.00 & $\mathbf{1.00}$ & 4.00 \\
& GAE & 1.83 & 2.50 & $\mathbf{1.67}$ & 4.00 \\
\midrule
\multirow[c]{2}{*}{ogbl-collab} & Node2Vec & 2.00 & 3.00 & $\mathbf{1.00}$ & 4.00 \\
& GAE & 2.00 & 3.00 & $\mathbf{1.00}$ & 4.00 \\
\midrule
\multirow[c]{2}{*}{ppi} & Node2Vec & 2.33 & 2.67 & $\mathbf{1.00}$ & 4.00 \\
& GAE & 2.89 & 2.06 & $\mathbf{1.11}$ & 3.94 \\
\bottomrule
\end{tabularx}

\end{table}

%% file: tables/alignment_loss_weight_comparison.tex
\begin{table}[ht]
    \centering
        \caption{
Comparison of $\alpha$-based \texttt{RAFEN} with simplified/weighted versions. We present the percentage difference of averaged link prediction AUC metric on the last snapshot over 25 runs. Values below zero denote better performance of the simplified/weighted variant, and values above zero denote better $\alpha$ variant performance. We mark outlier values with \textbf{bold}.
    }
    \label{tab:rafen_models_comparison}
\scalebox{0.90}{
\begin{tabular}{llc|cccccc}
\toprule
\textbf{Model} & \textbf{Selector} & \texttt{RAFEN} \textbf{variant} & \makecell[c]{\textbf{bitcoin} \\ \textbf{alpha}} & \makecell[c]{\textbf{bitcoin} \\ \textbf{otc}} & \makecell[c]{\textbf{fb} \\ \textbf{forum}} & \makecell[c]{\textbf{fb} \\ \textbf{messages}} & \makecell[c]{\textbf{ogbl} \\ \textbf{collab}}  & \makecell[c]{\textbf{ppi}} \\
\toprule
\multirow[c]{3}{*}{Node2Vec } & All & \texttt{$\alpha$ vs Simplified} 
& -0.55\% & 0.42\% & -0.58\% & 1.71\% & 0.59\% & 0.16\% \\
 & TB &
\texttt{REF\_$\alpha$ vs Weighted} 
& \textbf{4.84\%} & 0.00\% & \textbf{-2.59\%} & 1.34\% & 0.35\% & -0.49\% \\
 & EJ &
\texttt{REF\_$\alpha$ vs Weighted} 
& \textbf{5.01\%} & -0.42\% & 1.04\% & \textbf{2.79\%} & 0.47\% & -0.49\% \\
 \midrule
\multirow[c]{3}{*}{GAE} & All &
\texttt{$\alpha$ vs Simplified} &
0.49\% & 1.16\% & -0.12\% & -0.54\% & 0.76\% & 0.50\% \\
 & TB &
\texttt{REF\_$\alpha$ vs Weighted} 
& \textbf{2.07\%} & 0.91\% & -1.53\% & \textbf{2.12\%} & 0.25\% & 0.17\% \\
 & EJ & 
\texttt{REF\_$\alpha$ vs Weighted}
& \textbf{3.96\%} & 0.52\% & 0.81\% & -1.64\% & 1.27\% & 0.00\% \\
\bottomrule
\end{tabular}
}

\end{table}

%% file: tables/method_results.tex
\begin{sidewaystable}
    \centering
    \caption{Last snapshot link prediction evaluation results. We report AUC metric values with mean and standard derivation over 25 model retrains. We mark the top three results with \textbf{bold}. Methods mean ranks represent the ranking established upon mean AUC values over retrains.}
\label{tab:baselines_comparison}

\begin{tabular}{l|l|cccccc|c}
\toprule
 \textbf{Model} & \textbf{Alignment method}  &  \makecell[c]{\textbf{bitcoin} \\ \textbf{alpha}} & \makecell[c]{\textbf{bitcoin} \\ \textbf{otc}} & \makecell[c]{\textbf{fb} \\ \textbf{forum}} & \makecell[c]{\textbf{fb} \\ \textbf{messages}} & \makecell[c]{\textbf{ogbl} \\ \textbf{collab}} & \makecell[c]{\textbf{ppi}} & \textbf{Mean rank} \\
 \midrule
\multirow[c]{8}{*}{\rotatebox{90}{Node2Vec}}& (whole graph) & $70.60 \pm 5.80$ & $71.70 \pm 2.60$ & $\mathbf{91.40 \pm 0.80}$ & $71.70 \pm 4.70$ & $\mathbf{89.00 \pm 0.10}$ & $\mathbf{63.80 \pm 0.20}$ & \textbf{7.00} \\
 & (last snapshot only) & $72.70 \pm 4.40$ & $69.30 \pm 2.70$ & $80.00 \pm 2.40$ & $74.30 \pm 3.90$ & $84.50 \pm 0.10$ & $60.30 \pm 0.40$ & 10.42 \\
 \cmidrule{2-8}
 & Posthoc-PA & $74.60 \pm 7.00$ & $69.70 \pm 2.20$ & $84.40 \pm 2.10$ & $73.60 \pm 6.50$ & $84.50 \pm 0.20$ & $60.30 \pm 0.30$ & 10.00 \\
 & Posthoc-EJ & $76.50 \pm 7.50$ & $69.60 \pm 1.80$ & $84.20 \pm 2.10$ & $73.30 \pm 6.00$ & $84.50 \pm 0.20$ & $60.30 \pm 0.30$ & 10.33 \\
 & Posthoc-TB & $72.70 \pm 7.60$ & $70.20 \pm 2.80$ & $84.10 \pm 1.90$ & $74.00 \pm 5.40$ & $84.50 \pm 0.20$ & $60.30 \pm 0.30$ & 10.00 \\
 \cmidrule{2-8}
 & \makecell[l]{\texttt{RAFEN\_ALL}} & $73.10 \pm 5.70$ & $70.80 \pm 2.50$ & $\mathbf{87.20 \pm 1.70}$ & $74.60 \pm 5.40$ & $\mathbf{84.80 \pm 0.10}$ & $\mathbf{61.90 \pm 0.40}$ & \textbf{5.75} \\
 & \texttt{RAFEN\_Weighted\_EJ} & $70.20 \pm 7.20$ & $71.10 \pm 2.60$ & $85.80 \pm 1.80$ & $73.20 \pm 2.90$ & $84.70 \pm 0.20$ & $\mathbf{61.90 \pm 0.40}$ & 9.17 \\
 & \texttt{RAFEN\_Weighted\_TB} & $70.80 \pm 8.80$ & $71.40 \pm 3.00$ & $\mathbf{87.00 \pm 1.70}$ & $73.70 \pm 3.80$ & $\mathbf{84.80 \pm 0.10}$ & $61.90 \pm 0.30$ & 7.25 \\
 \cmidrule{1-8}
\multirow[c]{8}{*}{\rotatebox{90}{GAE}} & (whole graph) & $72.10 \pm 4.30$ & $\mathbf{77.80 \pm 1.70}$ & $68.20 \pm 6.30$ & $65.60 \pm 4.50$ & $80.70 \pm 0.20$ & $61.30 \pm 0.60$ & 10.00 \\
 & (last snapshot only) & $77.80 \pm 2.00$ & $74.50 \pm 1.90$ & $79.60 \pm 1.70$ & $\mathbf{75.80 \pm 3.80}$ & $78.20 \pm 0.30$ & $58.00 \pm 1.00$ & 9.92 \\
 \cmidrule{2-8}
 & Posthoc-PA & $\mathbf{82.70 \pm 2.80}$ & $74.40 \pm 2.30$ & $84.80 \pm 1.70$ & $\mathbf{75.70 \pm 3.80}$ & $78.20 \pm 0.30$ & $58.00 \pm 0.80$ & 8.42 \\
 & Posthoc-EJ & $\mathbf{82.60 \pm 3.00}$ & $74.50 \pm 2.20$ & $84.90 \pm 1.80$ & $75.40 \pm 3.30$ & $78.20 \pm 0.30$ & $58.10 \pm 0.70$ & 7.92 \\
 & Posthoc-TB & $\mathbf{82.30 \pm 3.50}$ & $74.80 \pm 2.00$ & $84.80 \pm 1.60$ & $\mathbf{75.90 \pm 3.50}$ & $78.20 \pm 0.30$ & $58.10 \pm 0.60$ & 7.58 \\
 \cmidrule{2-8}
 & \texttt{RAFEN\_ALL} & $81.50 \pm 2.60$ & $\mathbf{76.80 \pm 1.40}$ & $86.60 \pm 1.10$ & $74.00 \pm 3.50$ & $78.50 \pm 0.40$ & $59.30 \pm 0.50$ & \textbf{6.75} \\
 & \texttt{RAFEN\_Weighted\_EJ} & $80.10 \pm 2.60$ & $76.30 \pm 1.50$ & $86.10 \pm 1.20$ & $74.50 \pm 2.60$ & $78.00 \pm 0.30$ & $59.40 \pm 0.60$ & 8.00 \\
 & \texttt{RAFEN\_Weighted\_TB} & $80.50 \pm 3.30$ & $\mathbf{76.40 \pm 1.90}$ & $86.30 \pm 1.20$ & $73.90 \pm 2.80$ & $78.50 \pm 0.30$ & $59.30 \pm 0.50$ & 7.50 \\
\bottomrule 
\end{tabular}
\end{sidewaystable}